\documentclass[12pt]{article} 
\usepackage[russian,english]{babel}
\usepackage{changepage}
\usepackage{cite}
\usepackage[usenames]{xcolor}
\usepackage{comment}
\usepackage[shortlabels]{enumitem}
  \setlist[enumerate,1]{leftmargin=15pt}
\usepackage{float}\restylefloat{figure}
\usepackage[paper=a4paper]{geometry}
\usepackage{graphicx}
\usepackage{hyperref}
\usepackage{multicol}
\usepackage[leftmargin=16pt,vskip=6pt]{quoting}
\usepackage{wasysym}
\usepackage{xurl}
\newcommand\Ar{\medskip\noindent\texttt{A:\ }}

\renewcommand\phi{\varphi}
\newcommand{\Qn}{\medskip\noindent\texttt{Q:\ }}

\newcommand\ru[1]
  {\begin{otherlanguage}{russian}#1\end{otherlanguage}}

\title{On logic and generative AI}
\author{Yuri Gurevich and Andreas Blass\\
\normalsize University of Michigan}
\date{}

\begin{document}
\thispagestyle{empty}

\noindent
\textbf{\large Preamble}

\bigskip
This article was originally written for the June 2024 issue of the Bulletin of European Association for Theoretical Computer Science,
in the framework of the ``Logic in Computer Science'' column administered by Yuri Gurevich.
In the following pages, the article is reproduced as is.

\newpage\noindent
\textbf{\large Foreword by the columnist}\\[10pt]
The ongoing AI revolution raises many foundational problems.
For quite a while, I felt that the issue needs to be addressed in this column.
Not being an AI expert, I was looking for volunteers.
This didn't work, and so one day I took a deep breath and started to write an article myself.
Andreas Blass, my long-time collaborator, was reluctant to join me, but eventually he agreed.

A hundred years ago, logic was almost synonymous with foundational studies. 
I tried to rekindle that tradition in \cite{G247}.
The goal of the following dialog is to provoke young logicians with a taste for foundations to notice the foundational problems raised by the ongoing AI revolution.

\vspace{50pt}
\includegraphics[width=11cm]{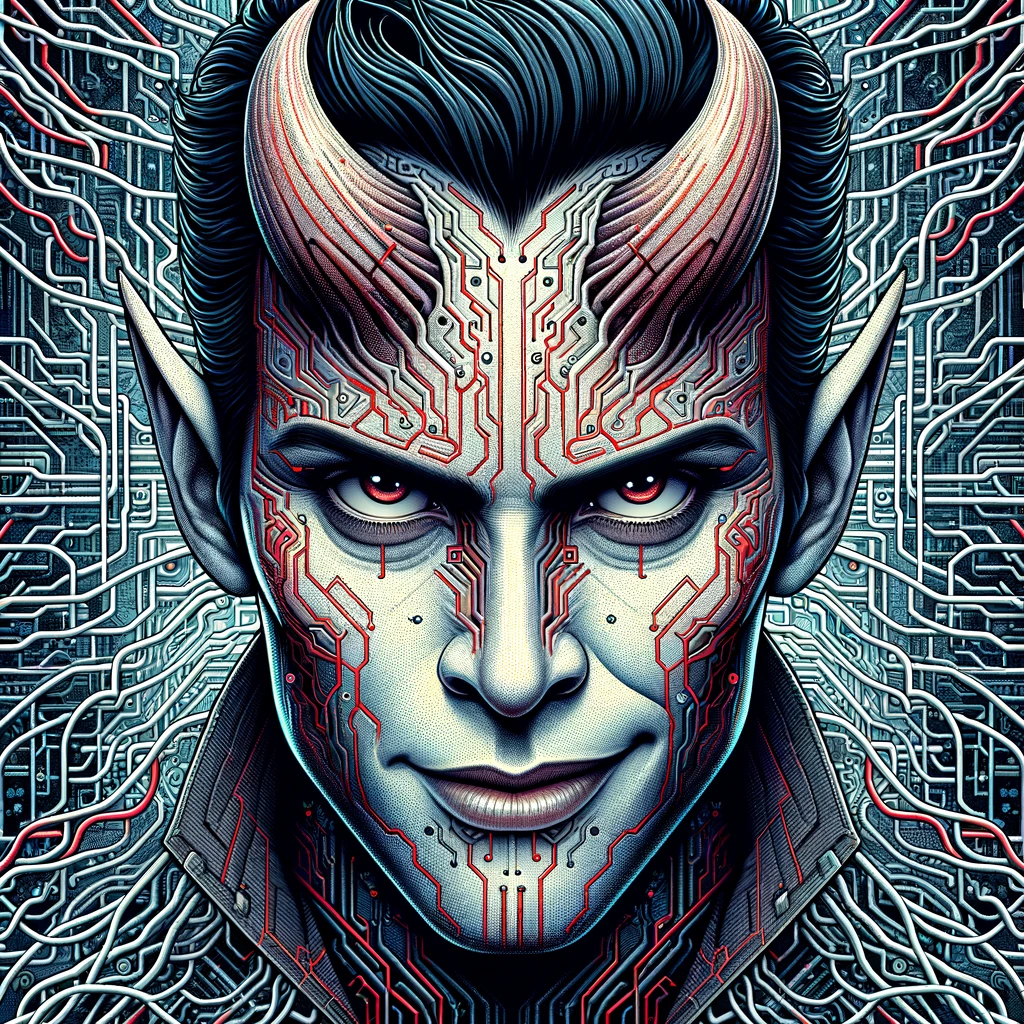}\\
\centering{\footnotesize
A whimsical picture of a neural net with a devil's head by the OpenAI tool Dall-E}

\newpage
\begin{center}
\textbf{\Large On logic and generative AI}\\[10pt]
{\large Yuri Gurevich and Andreas Blass\\[5pt] \normalsize 
University of Michigan in Ann Arbor, MI, USA}
\end{center}

\begin{quote}\raggedleft\small\it
I think the most beautiful thing about\\ deep learning is that it actually works.\\[1ex]
---Ilya Sutskever \cite[29:46]{Sutskever2020}
\end{quote}
\raggedright
\subsection{Thinking fast and slow\protect\footnotemark}
\footnotetext{The freewheeling conversations \cite{Altman, Gibson, LecunChat} are broken into ``chapters'' for the reader's convenience.
Following this example, our freewheeling dialog was broken into parts at the eleventh hour.\\
Below \texttt{Q} is Quisani, a former student of the first author, and \texttt{A} is the authors.}

\Qn I just learned that Daniel Kahneman, Nobel laureate in economics and the author of ``Thinking, fast and slow'' \cite{Kahneman}, passed away on March 27, 2024.
I heard a lot about this book but have never read it.
What did he mean by thinking fast and thinking  slow?

\Ar Daniel Kahneman and Amos Tversky discovered that human thinking is driven by two distinct systems, System~1 and System~2.

System 1 supports \emph{fast thinking}.  It
``operates automatically and quickly, with little or no effort and no sense of voluntary control \dots\
The capabilities of System~1 include innate skills that we share with other animals'' \cite[pp.~41-43]{Kahneman}.
System~1 is good at detecting patterns and reading situations on the fly. 
It allows us to make snap judgments and decisions without deliberation, but it is prone to biases and errors.

System 2 supports \emph{slow thinking} which is deliberate, analytical, and conscious. 
``System 2 allocates attention to the effortful mental activities that demand it, including complex computations. 
The operations of System~2 are often associated with the subjective experience of agency, choice, and concentration'' \cite[p~41]{Kahneman}.
System~2 can override the impulses and biases generated by System~1.

\Qn I probably rely on System~1 more than I should.

\Ar System~2 is much slower than System~1 and requires more effort.
``While walking comfortably with a friend, ask him to compute $23\times78$ in his head, and to do so immediately. He will almost certainly stop in his tracks ''\cite[p~74]{Kahneman}.
It is not surprising that often we tend to be lazy and rely on System~1 more than we should.

\Qn Have either of you met Kahneman?

\Ar Andreas: No, I haven't. \\
Yuri: Neither have I, even though, in the 1970s, when I was teaching at Ben Gurion University, I had a good chance to see Kahneman and Tversky at the Hebrew University of Jerusalem where they had their seminar in psychology. I was at the Hebrew University  on Wednesdays.
My Jerusalem colleagues spoke enthusiastically about the psychology seminar, and I intended to drop in some day, but never did. 
Those Wednesdays were too  busy for me: model theory seminar, set theory seminar, and work with the renowned logician Saharon Shelah.

\subsection{Is generative AI intelligent?}

\Qn Generative AI is getting a lot of press recently. 
Is it intelligent?

\Ar The AI revolution is impressive. 
Generative AI is used in many domains.
In particular, it is used to improve existing logic tools.
Generative AI is changing our lives.
But is it already intelligent? 
The issue is controversial and super interesting; it richly deserves a separate conversation. 

\Qn May we address it anyway, at least briefly?

\Ar Sure. So the thesis under discussion is that the current generative AI is intelligent.
Let's first hear skeptics.

Edward Gibson is a psycholinguistics professor at MIT and the head of the MIT Language Lab. 
In his recent interview with Lex Fridman, Gibson argues that the current LLMs are all about the form, not meaning.
\begin{quoting}
``I would argue they’re doing the form. 
They’re doing the form, they’re doing it really, really well. 
And are they doing the meaning? No, probably not. 
There’s lots of these examples from various groups showing that they can be tricked in all kinds of ways. 
They really don’t understand the meaning of what’s going on. 
And so there’s a lot of examples \dots\ which show they don’t really understand what’s going on'' \cite[01:34:33]{Gibson}.
\end{quoting}

\Qn Presumably, GPT-4 knows all grammatical rules of English in all existing grammar books.

\Ar The rules of natural languages may have statistical character and may not be written anywhere. 

\Qn Give me an example.

\Ar A typical American pronounces the one vowel in ``Blass'' as in ``glass'' and stresses the first syllable in ``Gurevich.'' 
In fact, ``Blass'' is a German name, and German doesn't have the English ``glass'' vowel. 
Similarly, ``Gurevich'' is a Russian name, and Russians stress the second syllable.

\Qn This is about pronunciation. 
Give me a text-based example.

\Ar One has \textbf{a} cold, \textbf{the} flu, or influenza. 

\Qn Wow, I am glad that I don't have to teach grammar.

\Ar Also, in America, one is ``in the hospital'' while, in England, one is ``in hospital.'' 
In America, the government ``proposes'' a law, while, in England, the government ``propose'' a law.

\Qn You mentioned ``skeptics.'' 
Whom else did you have in mind?

\Ar Yann LeCun, along with Yoshua Bengio and Geoffrey Hinton, received the 2018 Turing Award for their work on deep learning.
They are sometimes referred to as the ``Godfathers of Deep Learning.''
LeCun says that LLMs, large language models like GPT-4, aren't truly intelligent yet (and cannot take us to superhuman intelligence) because they essentially lack important capabilities \cite[00:02:47]{LecunChat}: 
\begin{enumerate}
\item understanding the physical world, 
\item  persistent memory,
\item reasoning, and
\item planning. 
\end{enumerate}
``That is not to say,'' adds LeCun, ``that autoregressive LLMs are not useful, they’re certainly useful; or that they’re not interesting; or that we can’t build a whole ecosystem of applications around them, of course we can. 
But as a path towards human-level intelligence, they’re missing essential components.'' 

At this point, we heard two skeptics. What do you think?

\Qn Gibson's argument did not convince me too much. 
I am thinking about clever extraterrestrials analysing the wet chemistry of our nervous system and wondering do humans really understand the meaning of what’s going on? 
There are lots of ways humans can be tricked into irrational behaviour.

LeCun doesn't look to me like a true skeptic.
He points out various ways to improve the current generative AI.

Also, there is a matter of definitions, in particular the definition of intelligence. 
Maybe we should be speaking about degrees of intelligence.

\Ar OK, let's turn our attention to the defence of the thesis. 
In a recent talk, Geoffrey Hinton made a strong case for the intelligence of the current large language models, LLMs.

\begin{quoting}
``They [LLMs] turn words into features, they make these features interact, and from those feature interactions they predict the features of the next word.
And what I want to claim is that these millions of features and billions of interactions between features that they learn, are understanding \quad \dots

This is the best model we have of how we understand.
So it's not like there's this weird way of understanding that these AI systems are doing and then this [is] how the brain does it.
The best that we have, of how the brain does it, is by assigning
features to words and having features [and] interactions'' \cite[0:14--15]{Hinton}.
\end{quoting}

\Qn But these allegedly intelligent LLMs hallucinate.

\Ar So do we. ``Anybody who's studied memory, going back to Bartlett in the 1930s, knows that people are actually just like these large language models.
They just invent stuff and for us, there's no hard line between a true memory and a false memory'' \cite[0:16]{Hinton}.

\Qn Interesting. I intend to listen to that talk of Hinton.

\Ar Another champion of the thesis is Ilya Sutskever, Chief Scientist at Open AI \cite{Sutskever2020,Sutskever2023}.
 
A brief discussion cannot do justice to an issue as complicated as the thesis. 
Let's move on.

\Qn Wait, there is a closely related issue: 
What about the future? 
Many AI experts say that generative AI is getting more intelligent than humans and may become a danger to us.

\Ar The chances are that its intelligence will be incomparable to ours.
Think of airplanes versus birds. 
Airplanes fly faster, but birds can land and take off almost anywhere. 
On the other hand, AI develops fast and we don't know the directions.
``It's tough to make predictions, especially about the future,'' said 
the baseball philosopher Yogi Berra.

In any case, you are right. 
Numerous AI experts worry that generative AI may become a danger to humans.
They point out that it may be used by bad actors for manipulating
electorates and waging wars. 

\begin{quoting}
``There are always misguided or ill-intentioned people,'' says Bengio, ``so it seems highly probable that at least one organisation or person would --- intentionally or not --- misuse a powerful tool once it became widely available'' \cite{Bengio}.
\end{quoting}

In the talk \cite{Hinton}, Hinton describes various scenarios he is worried about.
\begin{quoting}
``But the threat I'm really worried about,'' says Hinton, ``is the long term existential threat.
That is the threat that these things could wipe out humanity \dots

[W]hat happens if superintelligences compete with each other?\ \dots\
The one that can grab the most resources will become the smartest.
As soon as they get any sense of self-preservation, then you'll get
evolution occurring.
The ones with more sense of self-preservation will win and the more
aggressive ones will win''\ \cite[0:21--24]{Hinton}.
\end{quoting}

\Qn This is frightening indeed.

\subsection{Reasoning and logic in AI}

\Qn I am fascinated by LeCun's analysis above. 
Let's look at it more carefully.
Persistent memory sounds technical, not as profound as the three other capabilities.

\Ar Persistent memory would help with long-term coherence. 
If you heavily interact with an LLM for a long time, it would be highly desirable that the LLM maintains coherence and consistency.
This is a challenge for the current systems. 

\Qn This is a challenge for some humans as well \smiley{}

\Ar Also, in humans, persistent memory is vital for incubation of ideas, cross-pollination of ideas from different domains, and reasoning by analogy.

\Qn Concerning LeCun's item~3, what kind of reasoning is AI incapable of?

\Ar System~2 reasoning. 
An LLM  is basically just two files. One is a huge data file that reflects the information the model was trained on.
The other is an algorithm,  typically succinct, which usually embodies the model architecture as well as a probabilistic inference mechanism.
That mechanism works with a sequence of words. 
It starts with a given query, and runs in rounds. During one round, it infers another word and appends it to the current sequence.
(More exactly, it works with subword tokens which are not necessarily full words.)
  
Sometimes the next word is obvious, but sometimes it is very hard to figure out an appropriate next word. 
But the LLM cannot stop and think. 
In the AI parlance, ``it is allocating approximately the same amount of compute for each token it generates'' \cite[00:59:51]{Altman}.
In that sense it uses only fast thinking (System~1).
As far as we know, it is an open problem how to improve the process by incorporating slow, deliberate thinking (System~2).

\Qn Understanding of the real world seems to involve reasoning as well, both System~1 reasoning and System~2 reasoning, from evading predators (fast thinking) to quantum physics (slow thinking).
Thus reasoning is ubiquitous in AI, and AI needs many different kinds of reasoning. 
It would be natural if the role of logic in AI would grow and grow. 
And yet, Yuri wrote that ``the golden age of logic in artificial intelligence is behind us" \cite[\S4]{G247}. How come?

\Ar At the early stage of AI, the logic approach was dominating. 
A lot of good work was done; see the article ``Logic-based artificial intelligence'' in the Stanford Encyclopedia of Philosophy  \cite{Thomason}.
But then an important competitor arose.
Let us quote LeCun on the issue\footnotemark.
\footnotetext{The French original:
``Dans les ann\'ees 1950, tandis que les h\'eraults de l’intelligence
artificielle classique, bas\'ee sur la logique et l’exploration arborescente, en repoussent les limites, les pionniers de l’apprentissage commencent \`a donner de la voix. 
Ils d\'efendent l’id\'ee que, si l’on veut rendre les syst\`emes informatiques capables de t\^aches complexes, \`a la mani\`ere de l’animal et de l’homme, la logique ne suffit pas. 
Il faut se rapprocher du fonctionnement du cerveau, et donc rendre les syst\`emes capables de se programmer eux-m\^emes, en s’inspirant de ses m\'ecanismes d’apprentissage.
Cette partie de la recherche fond\'ee sur l’apprentissage profond (deep learning) et les r\'eseaux de neurones (artificiels) est celle \`a la\-quelle je me consacre. Elle est \`a l’\oe uvre dans toutes les applications spectaculaires actuelles, \`a commencer par la voiture autonome''\ \cite[p.~23]{LecunBook}.}
\begin{quoting}
``In the 1950s, while the heralds of classical AI, based on logic and tree-based exploration, were pushing back its limits, the pioneers of learning started to make their voices heard. 
They defended the idea that, if we want to make computer systems capable, like animals and humans,  of complex tasks, logic is not enough. 
We must get closer to the functioning of the brain, and therefore make the systems capable of self-programming, drawing inspiration from their learning mechanisms.
This part of research based on deep learning and (artificial) neural networks is the one to which I am devoted here. 
It is at work in all of today’s spectacular applications, starting with the autonomous car''\ \cite[p.~23]{LecunBook}.
\end{quoting}

\subsection{Real-world understanding}

\Qn Coming back to real-world understanding, robots need to navigate around obstacles and handle objects without causing damage to those objects, to themselves, and of course to people. 
This requires understanding of the physical world.
But, unlike robots, LLMs don't walk around and they have read much of what there is to read.
Shouldn't all that knowledge give them a near perfect understanding of the physical world?

\Ar Not really. 

\Qn Can you illustrate that?

\Ar Sure. Henri Lebesgue had an idea that we can build upon: there are cases where arithmetic does not apply.
``Suppose I put two animals in a cage and then two more. How many
animals will the cage then contain? \cite[p.~14]{Lebesgue}'' 
Let's present a version of this scenario to Google's Gemini.

\begin{enumerate}\small
\item[---] Suppose that I put 2 animals of one kind and 3 animals of another kind into a closed space. 
    How many animals will be there altogether?
\item[---] Adding the number of animals of each kind together, you will have 2 + 3 = 5 animals altogether.
\end{enumerate}

\Qn This is ridiculous. 
Gemini doesn't realize that the result depends on the kinds of animals. 
For example, the two animals of the first kind may consume some or all of the three animals of the other kind.
Maybe, Gemini takes into account only the moment that the you close the door of the closed space.

\Ar OK, let's check your hypothesis.
\begin{enumerate}\small
\item[---] What if I check the total number of animals not immediately but in one hour, one day, or one month?
\item[---] The total number of animals (5) wouldn't change regardless of whether you check on them in one hour, one day, or even one month. As long as no animals enter or leave the closed space, the initial sum of 2 and 3 will remain the same.
\end{enumerate}

\Qn One month is not a trivial period. 
Even if none of our original animals is a predator, some of the animals may die or have young.

\Ar  Gemini doesn't have your real-world experience\footnotemark.

\footnotetext{The chat with Gemini occurred on April~29th.
Long before that, we presented to Gemini another scenario (a coin tossed down from a tall building) that also exposed Gemini's lack of real-world knowledge.
In early May, we again presented both scenarios to Gemini and discovered that Gemini improved in the meantime. 
Despite the improvements, Gemini still showed, albeit in a less dramatic way, a lack of real-world knowledge.}

\Qn Have you seen imperfections with other LLMs?

\Ar In \S3, we mentioned an article on logic-based AI in the Stanford Encyclopedia of Philosophy (SEP). 
Before turning to SEP, we asked GPT-4 to suggest a good survey article on logic-based AI. 
GPT-4 hallucinated a combination of an existing title and well-known  authors who could write an article with this title but didn't.

\Qn This harks back to the issue that an LLM can't stop and think. 

\Ar In this particular case, the necessary thinking is algorithmic: just check the known sources.
So this hallucination seems to be a bug.

\subsection{Moravec's paradox}

\Qn It is ironic that Gemini can do impressive intellectual work but misses obvious things about the physical world.

\Ar You rediscovered Moravec's paradox from the 1980s.
Here's an instructive quote from Hans Moravec's 1988 book \cite[pp.~15--16]{Moravec}:
\begin{quoting}
``It seemed to me that, in the early 1970s, some of the creators of
successful reasoning programs suspected that the poor performance in the robotics work somehow reflected the intellectual abilities of
those attracted to that side of the research. 
Such intellectual snobbery is not unheard of, for instance between theorists and experimentalists in physics. 
But as the number of demonstrations has mounted, it has become clear that it is comparatively easy to make computers exhibit
adult-level performance in solving problems on intelligence tests or
playing checkers, and difficult or impossible to give them the skills of a one-year-old when it comes to perception and mobility.

In hindsight, this dichotomy is not surprising. 
Since the first multicelled animals appeared about a billion years ago, survival in the fierce competition over such limited resources as space, food, or mates has often been awarded to the animal that could most quickly produce a correct action from inconclusive perceptions. 
Encoded in the large, highly evolved sensory and motor portions of the human brain is a billion years of experience about the nature of the world and how to survive in it. 
The deliberate process we call reasoning is, I believe, the thinnest veneer of human thought, effective only because it is supported by this much older and much more powerful, though usually unconscious, sensorimotor knowledge. 
We are all prodigious olympians in perceptual and motor areas, so good that we make the difficult look easy. 
Abstract thought, though, is a new trick, perhaps less than 100 thousand years old. 
We have not yet mastered it. 
It is not all that intrinsically difficult; it just seems so when we do it.'' 
\end{quoting}
 
\Qn I don't buy that slow thinking is not all that intrinsically difficult.
He also seems to suggest that --- given a chance and time --- evolution will make slow thinking more efficient. 
Taking into account how wayward evolution is, the time in question may be humongous. 
But I digress.

\Ar In the Lex Fridman podcast, LeCun further develops Moravec's argument.
\begin{quoting}
``Those LLMs are trained on enormous amounts of texts, basically, the entirety of all publicly available texts on the internet, right? That’s typically on the order of $10^{13}$ tokens. Each token is typically two bytes, so that’s $2\cdot10^{13}$ bytes as training data. It would take you or me 170,000 years to just read through this at eight hours a day. So it seems like an enormous amount of knowledge that those systems can accumulate, but then you realize it’s really not that much data. If you talk to developmental psychologists, they tell you [that] a four-year-old has been awake for 16,000 hours in his or her life, and the amount of information that has reached the visual cortex of that child in four years is about $10^{15}$ bytes\ \cite[00:04:08]{LecunChat} \dots 

What that tells you is that through sensory input we see a lot more information than we do through language and that, despite our intuition, most of what we learn and most of our knowledge is through our observation and interaction with the real world, not through language''\ \cite[00:05:12]{LecunChat}.
\end{quoting}

\subsection{Thinking fast}

\Qn All this makes the fast-thinking grasp of the real world even more interesting and relevant. 
Clearly, LLMs would need such a grasp.

\Ar  They need it already. 
Have you recently spoken to an LLM-powered chatbot?

\Qn I try to avoid them and talk to humans.

\Ar This will be harder and harder to do. 

\Qn It is already plenty hard and sometimes virtually impossible; try to talk to OpenAI support.

But, OK, let's suppose that I am chatting with a bot representative of some company.

\Ar Imagine that they sent you something, say a laptop, and it arrived damaged, that you waited for that laptop longer than promised, and that you need one right now.
A human  would quickly realize that you are upset and would try to be extra empathetic. So should the bot. 

\Qn Yes, and this requires reading  the situation on the fly, fast thinking.

Now that I have some idea about fast thinking, I understand Yuri's question what the logic of fast thinking is in the February 2021 column.
But is there is a logic of fast thinking?

\Ar Much depends on what we mean by logic.
George Boole discovered an algebra of universal ``laws of thought'' \cite{Boole}, that is, universal laws of slow thinking. 
Is there a useful algebra of universal laws of fast thinking?
We doubt it. 
Fast thinking is rather far from exact reasoning.

\Qn Are there any meaningful rules of fast thinking?

\Ar Oh, yes. 
For example, we intuitively judge the frequency of an event by how easily examples come to mind, which can be influenced by recent exposure and emotions \cite[p.~244]{Kahneman}.
Such rules of thumb help us, given inconclusive perceptions, to produce quickly an action that is often beneficial but may be detrimental. 

Arguably logic should study the laws of thinking, including fast thinking, whatever they are. 
The intention, in the question about the logic of fast thinking, was to provoke young logicians with a taste for foundations to notice the nascent study of fast thinking. 

\Qn Studying fast thinking may expand logic tremendously.
Do you know historical examples where logic expanded so greatly?

\subsection{The dawn and heyday of logic}

\Ar One example is related to the emergence of logic as a separate discipline in the classical Greek period.

\Qn Was there any logic before that?

\Ar Various forms of structured argumentation were used before the classical Greek period, certainly in China and India.
But it was in Greece that logic became a separate discipline.

\Qn Does this have something to do with Athenian democracy?

\Ar The practice of Greek democracy wasn't limited to Athens, though Athens is the best-documented case.
The Athenian democracy was characterized by the participation of its (free male) citizens in the Assembly and courts.
There were many occasions where one had to convince numerous citizens rather than talking to gods or to a few aristocrats.
The necessity of convincing argumentation drove the development of logic as well as of rhetoric and demagoguery.
The contributions of Aristotle and his school were particularly significant in the birth of logic as a discipline. 
Aristotle's system of logic was virtually unchallenged until the 19th century.

\Qn Mathematics greatly advanced in the period from Aristotle to the 19th century. 
Mathematical arguments went beyond Aristotelian syllogisms.
They must have used much more sophisticated logic, and logic seems to precede mathematics, logically speaking \smiley{} 

\Ar Yes, already in the time of Aristotle mathematics involved logic beyond Aristotelian syllogisms. 
But for a long time mathematicians used deduction principles implicitly. 
After all, those principles are essentially trivialities. 
One learned them in the course of learning mathematics.

\Qn Still, there were many developments in mathematics, between Euclid's time and 19th century.

\Ar The most dramatic development was the invention of calculus.
Its use of infinitesimals and related notions, like infinite sums, certainly raised new logical issues.
Reasonable looking computations could contradict each other.

\Qn Give me an example.

\Ar Consider the infinite series $1 -1 + 1 - 1 + 1 -1 + \dots$
One way to sum it is $(1- 1) + (1 - 1) + (1 - 1) + \dots = 0$.
Another way is $1 + (-1 + 1) + (-1 + 1) + \dots = 1$.
Which way, if any, is the correct one?
The foundations of mathematics became problematic.

\Qn Today we know that the series does not converge.

\Ar It took a while for mathematicians to work out notions like convergence. 
The problem was actual infinity. 
Mathematicians used to work with finite objects and potential infinity, never actual infinity. 
E.g., for Euclid, the only lines were finite line segments.
 
\Qn In my experience, people mention infinitesimals but don't use them.

\Ar Eventually, the language of infinitesimals was replaced by the precise language of epsilon and delta, and thus the foundations of analysis got rid of the fuzziness induced by infinitesimals.

The introduction of set theory by Georg Cantor improved the situation further.
Cantor bravely dealt with actual infinity and made great progress.
Mathematicians quickly grew comfortable with set theory.
In particular, it was observed that ordered pairs can be coded as sets, which allowed modern set-theoretic definitions of notions like \emph{function} and \emph{relation} purely in terms of set membership. 

But at the same time paradoxes of set theory were discovered, and a real crisis arose in the foundations of mathematics.
It wasn't clear anymore which convincing reasoning is valid and which is not.

\Qn And, I guess, all this provoked the birth of mathematical logic.

\Ar Even before the discovery of paradoxes, important foundational work was done, notably by George Boole, Giuseppe Peano, and Gottlob Frege. 
But the foundational crisis put mathematical logic in the center of attention. 
The crisis was resolved, for all practical purposes, by the development of axiomatic set theory.

\Qn Did this make logic popular?

\Ar It surely did. Andrei Kolmogorov and John von~Neumann started as logicians. Alan Turing wrote his PhD thesis in logic \cite{Turing1938}. 
The universal Turing machine suggested to von~Neumann the idea of the universal electronic computer which made logic even more popular.

In 1956, when the famous Dartmouth workshop \cite{WikC} started AI as a field,
logic was at the zenith of its popularity. 
This is a partial explanation for the dominance of the logic approach at the early stage of AI.
By the way, the term ``artificial intelligence'' was coined by 
John McCarthy, Marvin Minsky, Nathaniel Rochester and Claude E.\ Shannon in their proposal for the Dartmouth conference.

\subsection{Logic in the age of AI}

\Qn In both cases, the birth of logic and the resolution of the foundational crisis in mathematics, there was a need, and logic rose to the occasion.
Demand preceded supply, so to speak.  
Do you think that, this time around, generative AI provides sufficient demand? 

\Ar It does. The demand is ubiquitous.

\Qn Give me an example.

\Ar A quick glance at a person gives you a surprising amount of information. 
We may learn the gender, estimate the age, etc.
Such rapid assessment was instrumental to our survival as a species.

\Qn It seems that, as in object-oriented programming, a person $P$ is viewed as an object with various attributes.

\Ar But the set of attributes may depend on the assessor; for example, a runner may see whether $P$ is a runner.
Also, the set of values that an attribute takes may be dynamic;
for example, a child may learn only whether $P$ is a child or an adult --- in contrast to an adult’s more precise estimate of $P$'s age.

\Qn Surely, psychologists and neuroscientists are already studying rapid assessment. What can logic add?

\Ar Rapid assessment is a case of fast thinking.
In \cite{G247}, Yuri asked what the logic of fast thinking is.
If and when laws of fast thinking are understood, at least to some extent, they may help us to analyze rapid assessment. 

\Qn Do you think that logic will rise to meet the AI challenge?

\Ar This is an interesting question. 
To us, logic is the study of reasoning, any kind of reasoning \cite{G247}.
There is no doubt that the kinds of reasoning which are relevant to AI will be studied. 

\Qn But will those studies be called logic? Maybe, AI logic?

\Ar At the moment this does not look probable. 
It is not predetermined what is or is not logic. 
Conflicting social processes are in play. 
Much depends on how --- and whether --- the logic community will address the challenge.

The story of infinitesimal calculus may be instructive.
See a sketch of that story in \cite[\S2]{G247}.
After the invention of infinitesimal calculus, a question arose how to reason with infinitesimals?
For a couple of centuries, mathematicians struggled with the problem and eventually largely solved it using the epsilon-delta approach that essentially eliminated infinitesimals.
In a sense, the mathematical heroes of that story worked as logicians but this wasn't recognized. 
This particular story has a happy, albeit somewhat bitter, end of sorts as far as logicians are concerned. 
Abraham Robinson came up with a proper logic of infinitesimals, which he called nonstandard analysis and which allows you to handle infinitesimals consistently \cite{Robinson}.
But, in the meantime, mathematicians got accustomed to the epsilon-delta approach and weren't too much interested in  nonstandard analysis which is arguably more elegant and efficient.

\Qn You wish of course that logic will be more successful in AI.

\Ar True.

\subsection*{Acknowledgments}
Many thanks to Naomi Gurevich, Ben Kuipers, Vladimir Lifschitz, and Moshe Vardi for useful comments.

\end{document}